\documentclass{article} %
\usepackage{iclr2026_conference,times}

\usepackage{amsmath,amsfonts,bm}

\def\eqref#1{equation~\ref{#1}}

\def\1{\bm{1}}

\DeclareMathAlphabet{\mathsfit}{\encodingdefault}{\sfdefault}{m}{sl}
\SetMathAlphabet{\mathsfit}{bold}{\encodingdefault}{\sfdefault}{bx}{n}

\usepackage{hyperref}
\usepackage{url}
\usepackage{graphicx}
\usepackage{enumitem}
\usepackage{xspace}
\usepackage{amsmath}
\usepackage{booktabs} %
\usepackage{geometry} %
\usepackage[linesnumbered,ruled,vlined]{algorithm2e}
\usepackage{colortbl}
\usepackage{multirow} 
\usepackage{booktabs}
\usepackage{array}
\usepackage{makecell}
\usepackage{wrapfig} 
\usepackage[table]{xcolor}
\usepackage{float}
\usepackage{authblk}
\definecolor{blue1}{RGB}{83,135,221}

\usepackage{mathtools} 
\usepackage{diagbox}
\usepackage{pifont}

\usepackage[most]{tcolorbox}
\tcbset{
  myexample/.style={
    colframe=blue1,
    fonttitle=\bfseries,
    left=.02in,
    right=.02in,
    bottom=.02in,
    top=.02in
  }
}

\newtcbtheorem[number within=section]{exmp}{Example}{myexample}{exmp}
\newtcbtheorem[number within=section]{reward}{Reward}{myexample}{exmp}
\newtcbtheorem[number within=section]{prompt}{Prompt}{myexample}{prompt}
\newtcbtheorem[number within=section]{prmprompt}{PRM Step Labeling Prompt}{myexample}{prmprompt}

\newcommand{\algoname}{GRPO-VPS}
\newcommand{\eg}{e.g.\@\xspace}

\title{\algoname{}: Enhancing Group Relative Policy Optimization with Verifiable Process Supervision for Effective Reasoning}

\author{
    Jingyi Wang$^1$\thanks{Equal contribution.}, Lei Zhu$^3$\protect\footnotemark[1], Tengjin Weng$^2$, Song-Li Wu$^1$, Haochen Tan$^3$, \\
    Jierun Chen$^3$, Chaofan Tao$^3$, Haoli Bai$^3$, Lu Hou$^3$, Lifeng Shang$^3$, 
    Xiao-Ping Zhang$^1$\thanks{Corresponding author.}
}
\affil[]{
    $^1$Tsinghua University \quad $^2$Shenzhen University \quad $^3$Huawei Noah's Ark Lab
}

\iclrfinalcopy %
\begin{document}

\maketitle
\begin{abstract}

Reinforcement Learning with Verifiable Rewards (RLVR) has advanced the reasoning capabilities of Large Language Models (LLMs) by leveraging direct outcome verification instead of learned reward models. Building on this paradigm, Group Relative Policy Optimization (GRPO) eliminates the need for critic models but suffers from indiscriminate credit assignment for intermediate steps, which limits its ability to identify effective reasoning strategies and incurs overthinking. In this work, we introduce a model-free and verifiable process supervision via probing the model’s belief in the correct answer throughout its reasoning trajectory. By segmenting the generation into discrete steps and tracking the conditional probability of the correct answer appended at each segment boundary, we efficiently compute interpretable segment-wise progress measurements to refine GRPO's trajectory-level feedback. This approach enables more targeted and sample-efficient policy updates, while avoiding the need for intermediate supervision derived from costly Monte Carlo rollouts or auxiliary models. Experiments on mathematical and general‑domain benchmarks show consistent gains over GRPO across diverse models: up to 2.6‑point accuracy improvements and 13.7\% reasoning‑length reductions on math tasks, and up to 2.4 points and 4\% on general‑domain tasks, demonstrating strong generalization.
\end{abstract}

\section{Introduction}

\begin{figure*}[t]
    \centering
    \resizebox{\textwidth}{!}{%
        \includegraphics{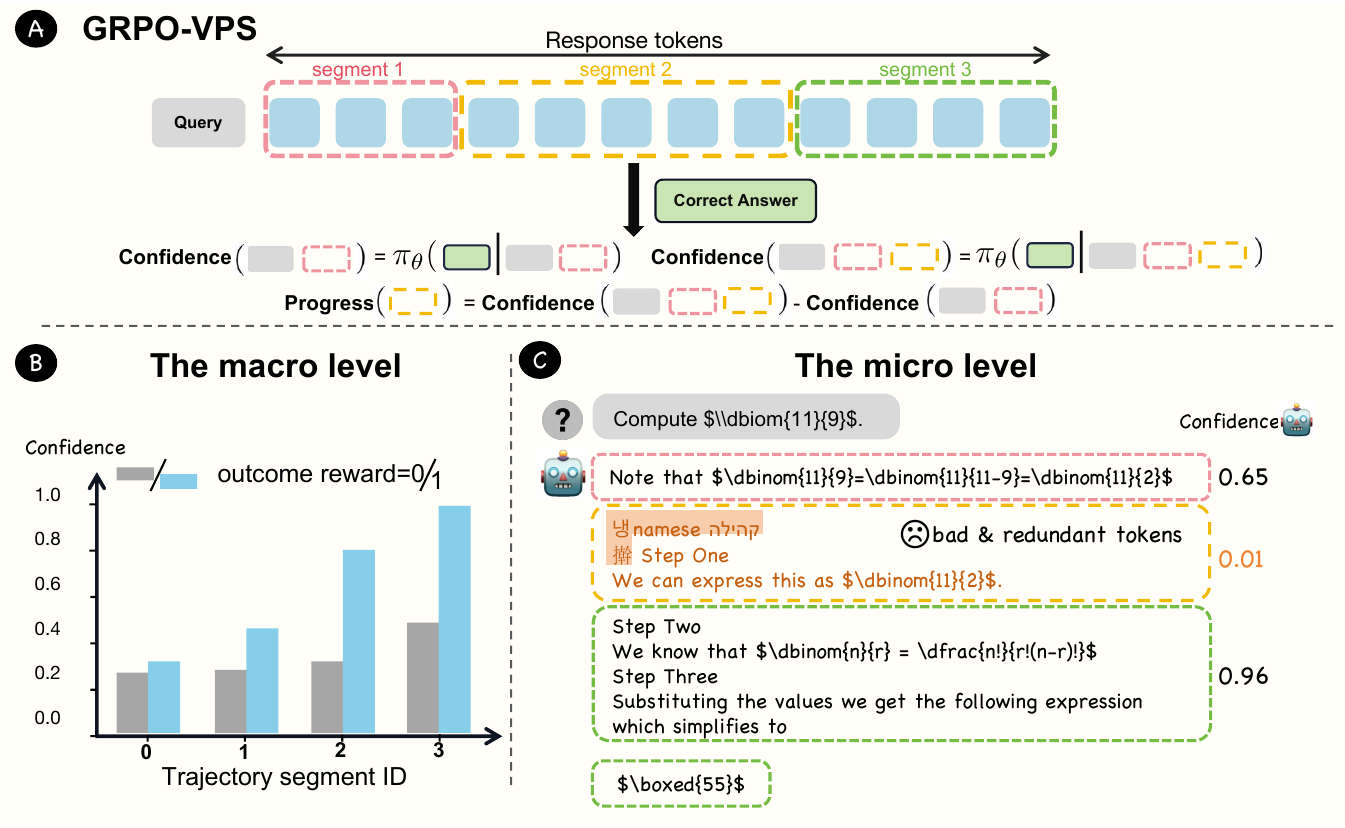}
    }
    \caption{\textbf{(A)}~\algoname{} supervises intermediate reasoning via a segment-wise process signal computed as the change in the model’s belief in the correct answer across consecutive reasoning segments.
    \textbf{(B)} At the macro level, we visualize how the probed confidence evolves in the reasoning models. 
    Trajectories that ultimately lead to correct answers exhibit more pronounced upward trends.
    \textbf{(C)} At the micro level, reasoning chunks with negative confidence increments often contain hallucinations, redundancies, or unhelpful detours.}
    \label{fig:pipline}
    \vspace{-2mm}
\end{figure*}

Advanced by Reinforcement Learning with Verifiable Rewards (RLVR)~\citep{shao2024deepseekmath, yu2025dapo}, Large Language Models (LLMs) have demonstrated remarkable capabilities in complex reasoning tasks, ranging from mathematical problem solving~\citep{OpenAI_2024_LearningReasonLLMs, team2025kimi, shao2024deepseekmath} to multi-hop question answering~\citep{huang2025rag, song2025r1}. The success of RLVR is largely attributed to computing rewards via direct outcome verification, rather than relying on reward models that complicate the training pipeline and are prone to reward hacking~\citep{yu2025dapo}. In a similar vein, Group Relative Policy Optimization (GRPO)~\citep{shao2024deepseekmath} eliminates critic models for token-level advantage estimation, instead uniformly propagating trajectory-level advantages to intermediate steps. While this simplification avoids the challenges of training a critic model and reduces associated overhead, indiscriminate credit assignment hinders sample efficiency and limits the policy model’s ability to learn effective reasoning strategies~\citep{qu2025optimizing}.

To address this limitation, we explore enhancing GRPO with model-free, verifiable process supervision derived from the annotated final answer. Our key insight is that the contribution of intermediate reasoning steps can be probed by the probability increment of the reference answer appended at corresponding breakpoints. This is supported by observations in Figure~\ref{fig:pipline}: (1) at the macro level, the average probed probability increases as reasoning progresses, with a more pronounced trend for trajectories that ultimately reach the correct answer; and (2) at the micro level, reasoning segments that reduce the probed probability tend to be of low quality. Based on these observations, our method leverages the model’s own reasoning trace to generate localized supervision signals, enabling more targeted and effective policy updates. Specifically, we segment the model’s response into discrete reasoning segments and strategically concatenate the correct final answer at each segment boundary. By extracting the model’s conditional probability of the correct answer at these positions, we obtain a proxy for its evolving belief state. The differences in these probabilities between adjacent segments serve as segment-wise supervision signals, complementing trajectory-level advantages and quantifying the contribution of each reasoning segment toward the final outcome.

This approach offers two key benefits: (1) it provides dense, interpretable feedback aligned with the model's internal decision flow and (2) it avoids reliance on auxiliary models~\citep{schulman2017proximal, zha2025rl, cui2025process, he_2024_16998085} or Monte Carlo rollouts~\citep{qu2025optimizing, dai2025s}, ensuring high efficiency and scalability and adhering to the design principles established by RLVR and GRPO. Through this fine-grained supervision mechanism, we aim to enhance the sample efficiency of RL training, paving the way for learning more effective and efficient reasoning behaviors.

Our experiments show substantial gains across four math reasoning benchmarks. Compared to GRPO, our method achieves up to +2.6 points Pass@1 on Qwen2.5-Math-1.5B and +1.1 point on Qwen2.5-Math-7B, while concurrently reducing reasoning length by 11.0\% to 13.7\%.  It also consistently outperforms the GRPO variant~\citep{dai2025s} that relies on costly Monte Carlo rollouts for segment-wise advantage estimation and auxiliary models~\citep{cui2025process, he_2024_16998085, schulman2017proximal}. Furthermore, evaluation on four general-domain reasoning benchmarks confirms strong generalization, with gains of 1.8 points on MMLUPro and 2.4 points on TheoremQA. These results highlight the effectiveness and scalability of our verifiable process supervision in delivering more accurate and concise reasoning.

In summary, our main contributions are:
\begin{itemize}[leftmargin=1.5em]
    \item We identify and empirically validate that a model’s evolving belief in the correct answer can serve as a model-free, interpretable signal for reasoning quality of intermediate steps. This enables fine-grained supervision without auxiliary models or Monte Carlo rollouts.
    \item We propose \algoname{}, a simple yet effective approach to enhance GRPO with granular, segment-wise process supervision, avoiding indiscriminate credit assignment and improving sample efficiency. %
    \item Our empirical results show that the method achieves strong performance on challenging math reasoning tasks, demonstrating that our method enhances both reasoning effectiveness and efficiency in comparison with GRPO and its variants.
\end{itemize}

\section{Related Work}

\textbf{Group Relative Policy Optimization (GRPO)}.
Reinforcement Learning with Verifiable Rewards (RLVR) has become a prominent paradigm for fine-tuning LLMs, using definitive signals from rule-based verifiers to circumvent the need for costly and potentially biased reward models \citep{shao2024deepseekmath, yu2025dapo}. Within this paradigm, GRPO~\citep{shao2024deepseekmath} offers a lightweight and efficient alternative to critic-based algorithms like PPO \citep{schulman2017proximal}. By comparing final outcomes across a group of sampled trajectories, GRPO eliminates the need for a separate value network. However, this simplification comes at the cost of indiscriminate credit assignment: a single, trajectory-level reward is uniformly propagated to all intermediate tokens. This can inadvertently reinforce spurious reasoning steps in a successful trajectory or penalize promising partial logic in a failed one. Our work addresses this limitation by introducing a fine-grained process supervision mechanism, enhancing its credit assignment capabilities without sacrificing its lightweight nature.

\vspace{.5em}
\noindent
\textbf{Process supervision for reasoning}. Recent work has explored injecting fine-grained supervision into the reasoning process to better guide long-form generation. These efforts can be broadly categorized into model-based and model-free approaches. Model-based supervision utilizes an auxiliary model to provide fine-grained feedback. Critic-based methods, often using PPO, train a value network to estimate the expected return from intermediate states \citep{yue2025vapo,kazemnejad2024vineppo}. However, in long-horizon reasoning tasks, the critic's signal can diminish or become unreliable due to the long delay in receiving the final outcome reward \citep{shao2024deepseekmath, yue2025vapo}. Process Reward Models (PRMs) \citep{lightman2023let, wang2023math} offer an alternative but are typically trained offline, making them vulnerable to reward hacking and distributional shift. These approaches introduce significant system complexity, requiring an extra model to be trained, maintained, and served alongside the policy.
Model-free process supervision aims to provide granular feedback without auxiliary models. Recent works have made progress in this direction. For instance, S-GRPO \citep{dai2025s} introduces a "serial group" objective with decaying rewards to encourage earlier, more efficient reasoning. MRT \citep{qu2025optimizing} frames the problem as 
meta reinforcement learning
and computes a dense ``progress'' reward based on the change in the likelihood of eventual success. Their reward estimation can require complex, rollout-based procedures or multiple generation branches from intermediate states, which compromises training efficiency. In contrast, our method simplifies the process supervision workflow by deriving a high-quality signal from the known ground-truth answer, requiring only a single forward pass per generated trajectory. This makes our approach more efficient while still providing the benefits of fine-grained, verifiable process feedback.

\section{Methodology: Process Supervisions from Verifiable Outcomes}
LLMs trained under GRPO still suffer from indiscriminate credit assignment, where sparse outcome-based rewards fail to guide intermediate reasoning steps. To address this, we present a verifiable process supervision framework for enhancing GRPO with fine-grained credit assignment. We first introduce a segmentation strategy that uses token-level entropy to identify high-uncertainty transitions and partition trajectories into semantically meaningful reasoning steps (Section~\ref{sec:Segmentation}). We then introduce segment-wise progress estimation to quantify the contribution of each reasoning segment based on changes in model confidence (Section~\ref{sec:progress}). Finally, we incorporate this localized feedback into GRPO’s token-level updates, forming a hybrid advantage that fuses outcome-based and process-level signals (Section~\ref{sec:advantage}).

\subsection{Reasoning Process Segmentation}
\label{sec:Segmentation}

Recent studies have revealed that performance gains in RLVR are primarily driven by critical decision points characterized by high token-level uncertainty~\citep{yang2025not, wang2025beyond}. Inspired by SPO~\citep{guo2025segment}, we adopt an Adaptive Entropy-based Cutpoint Partition strategy, leveraging token-level entropy to robustly identify reasoning "junctions" where the model's trajectory is likely to diverge.

Formally, given a response $o$ of length $T$, We identify a set of candidate cutpoints  $\mathcal{U} \subseteq \{1, \dots, T\}$ by selecting tokens whose entropy exceeds an adaptive threshold:
\begin{equation}
\mathcal{U} = \{ t \mid  e_t^{i} \geq \tau \},
\end{equation}
where $\tau$ is determined from the entropy distribution of $o$ (\eg via a perentile-based rule).
To partition $o$ into  $M$ reasoning segments $(z_{1}, \dots, z_{M})$, we choose boundary indices 
$\{t_1, \dots, t_{M+1}\}$ with $t_1=1$ and $t_{M+1}=T+1$, such that the number of cutpoints in each segment is approximately balanced:
\begin{equation}
|\mathcal{U} \cap [t_m, t_{m+1})| \approx \frac{|\mathcal{U}|}{M}, \quad \forall m \in \{1, \dots, M\}.
\end{equation}
This heuristic ensures that each segment contains a comparable number of high-entropy positions, yielding a balanced and semantically meaningful segmentation of the reasoning trajectory. Our experiments demonstrate that this adaptive strategy yields superior performance compared to fixed-token partition (see Section~\ref{abla}).

\subsection{Progress as Process Supervision}
\label{sec:progress}

Based on the reasoning process segmentation, we propose to leverage segment-wise progress as a form of process supervision to address the indiscriminate credit assignment of GRPO. This formulation provides a dense, model-free, and scalable supervision signal that quantifies the incremental contribution of each reasoning segment toward the correct final answer.

Given an input prompt $x$ and a trajectory $o$ generated by the policy $\pi_\theta$, we compute a segment-wise confidence score \( C(z_{\leq k}) \) representing the model's conditional probability of the target answer \( y^* \) after generating the first \( k \) reasoning steps:

\begin{equation}
C(z_{\leq k}) = \pi_\theta(y^* \mid x, z_{\leq k})
\end{equation}

where $x$ is the input question and $z_{\leq k} = (z_1, \ldots, z_k)$ denotes partial reasoning trace up to segment $k$. The initial value, before any reasoning, is $C_0 = P(y^* \mid x)$.

To quantify the contribution of each reasoning segment, we define a segment-wise progress score, denoted as \( \Delta C_k \), is then computed as the change in this confidence score, effectively isolating the contribution of that specific step:
\begin{align}
\Delta C_k &= C(z_{\leq k}) - C(z_{\leq k-1}) \notag \\
          &= \pi_\theta(y^* \mid x, z_{\leq k}) 
             - \pi_\theta(y^* \mid x, z_{\leq k-1}) \notag \\
\end{align}
where \( \Delta C_k \in [-1, 1] \) due to the probabilistic range of confidence scores. This results in a vector $\Delta C = [\Delta C_1, \ldots, \Delta C_m]$ of segment-level supervision signals for each trajectory. It reflects how much each segment improves (or worsens) the model's belief in the final answer.

\subsection{GRPO with Process Supervision}
\label{sec:advantage}

We design a hybrid advantage signal that fuses sparse outcome-level feedback with dense, segment-level process supervision. For each prompt, we sample a group of $G$ trajectories $\{o^1, o^2, \ldots, o^G\}$ from the policy $\pi_\theta$. Each trajectory \( o^i = (z_1^i, \ldots, z_m^i, y^i) \) with binary correctness label \( r^i \in \{0, 1\} \), we compute the group-relative advantage:
\begin{equation}
A^i=  r^i - \frac{1}{G}\sum_{j=1}^{G} r^j,
\end{equation}
where \( G \) is the number of responses sampled for the same prompt. To complement this global signal, we inject a localized feedback term \( \Delta C_k \), which quantifies the incremental gain in the model's belief in the correct answer after each reasoning segment \( z_k^i \), as defined in Section~\ref{sec:progress}. The final hybrid advantage at step \( k \) is:
\begin{equation}
\label{algori}
\tilde{A}_k^i = \underbrace{A^i}_{\text{Outcome}} + \underbrace{\alpha \cdot \Delta C_k}_{\text{Process}},
\end{equation}
where \( \alpha \) is a weighting factor balancing the two components. We empirically set $\alpha=1.2$ and found it work well. Sensitivity to $\alpha$ can be found in Appendix ~\ref{app:sensitive}.

We then define the final on-policy gradient estimator as:

\begin{equation}
\label{eq:final_grad_two_part}
\nabla_{\theta} J(\theta) = \frac{1}{G}\sum_{i=1}^{G} \sum_{k=1}^{M} \left( A^i + \alpha \cdot \Delta C_k \right) \cdot \nabla_{\theta} \log \pi_{\theta}(z_k^i \mid x, z_{<k}^i)
\end{equation}
where the total advantage combines two signals: $A^i$ provides sparse, trajectory-level feedback based on the final outcome, while $\alpha \cdot \Delta C_k$ injects dense, segment-level guidance reflecting the progress toward the correct answer. The full algorithm of \algoname{}~is shown in Algorithm~\ref{VPS_algori}.

\begin{algorithm}[t]
\caption{GRPO WITH VERIFIABLE PROCESS SUPERVISION}
\label{VPS_algori}
\KwIn{Question set $\mathcal{D}$; Base policy $\pi_{\theta_b}$; 
      Entropy percentile $p$; Segment number $m$; 
      Progress reward weight $\alpha$; Training steps $S$}
\KwOut{Updated policy parameters $\theta$}

Initialize policy $\pi_\theta \leftarrow \pi_{\theta_b}$ \\
\For{iteration $= 1, \dots, S$}{
    Sample a mini-batch $\mathcal{D}_b \subset \mathcal{D}$ \\
    \For{each question $x \in \mathcal{D}_b$ with target answer $y^*$}{
        Generate full trajectory $o  \sim \pi_\theta(\cdot|x)$ \\
        Compute token entropies $e_t$ \\
        Segment $o = (z_1,\dots,z_m, y)$ using adaptive entropy-based cutpoints \\
        Compute initial confidence $C_0 = \pi_\theta(y^* \mid x)$ \\
        \For{$k = 1$ to $m$}{
            $C_k = \pi_\theta(y^* \mid x, z_{\le k})$ \\
            $\Delta C_k  = C_k - C_{k-1}$ \\
        }
        Compute hybrid advantages $\tilde{A}_t$ using Eq.~\ref{algori} \\
        Update policy $\pi_\theta$ using policy gradient with $\tilde{A}_t$ \\
    }
}
\end{algorithm}

\section{Experiment}
\subsection{Setup}
\textbf{Models and baselines.} We conduct experiments on two model families, including Qwen2.5-Math-1.5B, Qwen2.5-Math-7B~\citep{yang2024qwen25mathtechnicalreportmathematical} and Gemma-2-2B-it~\citep{gemma_2024}. To ensure fair comparison, we include a comprehensive set of baselines categorized by their use of outcome-level vs. process-level supervision:
\begin{itemize}[leftmargin=1.5em]
    \item \textbf{Outcome Supervision Only.} This category includes methods that rely solely on final answer correctness for reward assignment. We consider GRPO and its recent variants, DrGRPO~\citep{liu2025understanding} and GSPO~\citep{zheng2025group}, which enhance group-wise comparison or propagate advantages with entropy-based mechanisms. We also include the BASE models without RL fine-tuning for reference.
    \item \textbf{With Process Supervision.} This group covers methods that incorporate intermediate supervision beyond outcome-level rewards. We evaluate S-GRPO~\citep{dai2025s}, which relies on  Monte Carlo rollouts with forced early stops to construct sub-trajectories, and assigns segment-level rewards based on their predicted outcomes. We also compare against PRIME-style reward modeling, represented by the public Eurus-2-7B-PRIME~\citep{cui2025process, yuan2024implicitprm}, and a controlled variant where we fine-tune Qwen2.5-Math-1.5B and 7B with Skywork-o1-prm using GRPO. These baselines provide strong comparisons for evaluating the effectiveness of verifiable process supervision in our method.
\end{itemize}

\noindent
\textbf{Training setup.} In line with prior work~\citep{liu2025understanding}, we use MATH~\citep{hendrycks2021measuring}, which contains 7,500 problems. We train the models using the verl framework~\citep{sheng2024hybridflow}. We sample 8 
rollouts per prompt, with a temperature of 1.0 and the maximum response length of 3,072 tokens. The batch size is set to 512, the mini-batch size to 128, and the learning rate to $1 \times 10^{-6}$. The training is conducted on a single node with 8 × H800 GPUs. More hyperparameter settings can be found in Appendix~\ref{appd:para}.

\noindent
\textbf{Evaluation setup.} We evaluate on four widely used math reasoning benchmarks, including the test sets of MATH, AIME 2024~\citep{AIME_2024_Feb}, AMC23~\citep{AMC12_2023_Nov} and OlympiadBench~\citep{he2024olympiadbench}. We set temperature to 1.0, top $p$ set to 1, and maximum output length set to 3,072 tokens for inference. Due to the high variance of the outputs from reasoning models, 
we report the average Pass@1 over 4 runs.
To ensure accurate evaluation, we utilize Math-Verify\footnote{\url{https://github.com/huggingface/Math-Verify}} to check for answer equivalence.

\begin{table*}[t]
\centering
\caption{Experimental results on Qwen-Math Models. BASE denotes the corresponding Qwen or Gemma base model without any fine-tuning.}
\label{tab:qwenmath_results}
\resizebox{\textwidth}{!}{%
\begin{tabular}{l|cc|cc|cc|cc|cc}
\toprule
\multirow{2}{*}{\textbf{Model}} & \multicolumn{2}{c|}{\textbf{AMC23}} & \multicolumn{2}{c|}{\textbf{AIME24}} & \multicolumn{2}{c|}{\textbf{MATH}} & \multicolumn{2}{c|}{\textbf{OLYMPIAD}}  & \multicolumn{2}{c}{\textbf{Overall}}\\
& \textbf{Pass@1} & \textbf{AvgToken} & \textbf{Pass@1} & \textbf{AvgToken} & \textbf{Pass@1} & \textbf{AvgToken} & \textbf{Pass@1} & \textbf{AvgToken} & \textbf{Pass@1} & \textbf{AvgToken}\\
\midrule

\rowcolor{gray!7} 
\multicolumn{11}{c}{\textbf{Qwen2.5-Math-1.5B}} \\
\midrule

\rowcolor{gray!7} 
\multicolumn{11}{l}{\textbf{Outcome Supervision Only}} \\
BASE                & 16.3   & 5534 & 5.8 & 5782 & 22.6 & 5008 & 19.0 & 4624 &$15.9_{\textcolor{blue}{\phantom{-00.0}}}$&$5237_{\textcolor{blue}{\phantom{-00.0\%}}}$\\

GRPO                & 46.9  & 3872 & 18.3 & 4655 & 68.8 & 2749 & 30.0 & 3811 &$41.0_{\textcolor{red}{+25.1}}$ & $3772_{\textcolor{blue}{-28.0\%}}$\\
DrGRPO              & 44.4  & 3681 & \textbf{20.0} & 4878 & 70.5 & 2757 & 30.8 & 3707 & $41.4_{\textcolor{red}{+25.5}}$& $3756_{\textcolor{blue}{-28.3\%}}$\\

GSPO                & 45.0     & 4044 & 13.3 & 5146 & 68.0 & 2778 & 28.9 & 3727 &$38.8_{\textcolor{red}{+22.9}}$& $3924_{\textcolor{blue}{-25.1\%}}$\\

\rowcolor{gray!7} 
\multicolumn{11}{l}{\textbf{With Process Supervision}} \\
GRPO w/ Skywork-1.5B & 50.0     & 3527 & 11.7 & 4738 & 71.4 & 2758 & 29.8 & 3403 &$40.7_{\textcolor{red}{+24.8}} $& $3607_{\textcolor{blue}{-31.1\%}}$ \\
S-GRPO              & 46.3   & 3535 & 14.2 & 4580 & 67.5 & 2664 & 28.3 & \textbf{3313} &$ 39.1_{\textcolor{red}{+23.2}}$& $3523_{\textcolor{blue}{-32.7\%}}$ \\
\rowcolor{cyan!20} 
\textbf{\algoname{}}          & \textbf{55.0}     & \textbf{3425} & 15.0 & \textbf{4278} & \textbf{72.2} & \textbf{2391} & \textbf{32.1} & 3326 &$\textbf{43.6}_{\textcolor{red}{+27.7}}$ &$\textbf{3355}_{\textcolor{blue}{-35.9\%}}$\\

\midrule
\rowcolor{gray!7} 
\multicolumn{11}{c}{\textbf{Qwen2.5-Math-7B}} \\
\midrule

\rowcolor{gray!7} 
\multicolumn{11}{l}{\textbf{Outcome Supervision Only}} \\
BASE                & 23.8   & 4485 & 10.8 & 5390 & 31.9 & 4370 & 18.8 & 4699 & $21.3_{\textcolor{blue}{\phantom{-00.0}}}$ &$4736_{\textcolor{blue}{\phantom{-00.0\%}}}$\\
GRPO                & 62.5    & 3552 & 30.0 & 4730 & 75.4 & 2671 & 36.8 & 3527 & $51.2_{\textcolor{red}{+29.9}}$ & $3620_{\textcolor{blue}{-23.6\%}}$\\

DrGRPO              & 64.4  & 3496 & 28.3 & 4449 & 75.5 & 2700 & 35.9 & 3490 & $51.0_{\textcolor{red}{+29.7}}$ & $3534_{\textcolor{blue}{-25.4\%}}$\\

GSPO                & 60.6  & 3725 & 30.8 & 4191 & 75.2 & 2661 & 36.8 & 3481 & $50.9_{\textcolor{red}{+29.6}}$ & $3515_{\textcolor{blue}{-25.8\%}}$\\

\rowcolor{gray!7} 
\multicolumn{11}{l}{\textbf{With Process Supervision}} \\
Eurus-2-7B-PRIME    & \textbf{65.0}     & 4368 & 15.0 & 5731 & \textbf{78.3} & 3024 & \textbf{42.2} & 4405 & $50.1_{\textcolor{red}{+28.8}}$ & $4382_{\textcolor{blue}{-\phantom{0}7.5\%}}$\\
GRPO w/ Skywork-7B   & 63.8  & 3990 & 30.0 & 4697 & 75.6 & 2736 & 38.8 & 3740 & $52.0_{\textcolor{red}{+30.7}}$ & $3791_{\textcolor{blue}{-20.0\%}}$ \\
S-GRPO              & 61.9  & 3235 & 25.8 & \textbf{3723} & 74.9 & 2464 & 35.8 & 3123 & $49.6_{\textcolor{red}{+28.3}}$ & $3136_{\textcolor{blue}{-33.8\%}}$ \\
\rowcolor{cyan!20} 
\textbf{\algoname{}}             & \textbf{64.8}     & \textbf{3220} & \textbf{31.7} & 3829 & 75.6 & \textbf{2372} & 37.2 & \textbf{3064} & $\textbf{52.3}_{\textcolor{red}{+31.0}}$ & $\textbf{3121}_{\textcolor{blue}{-34\%}}$\\

\midrule
\rowcolor{gray!7} 
\multicolumn{11}{c}{\textbf{Gemma-2-2B-it}} \\
\midrule

BASE                & 3.8   & \textbf{392} & 0.0 &  418 & 21.3 & 345 & 2.8 & 408 & $6.9_{\textcolor{blue}{\phantom{-00.0}}}$ &$391_{\textcolor{blue}{\phantom{-00.0\%}}}$\\
GRPO                & 7.5    & 3205 & 0.0 & 3183 & 31.0 & 2938 & 5.9 & 3133 & $11.1_{\textcolor{red}{+4.3}}$ & $3115_{\textcolor{blue}{+696\%}}$\\
\rowcolor{cyan!20} 
\textbf{\algoname{}}                & \textbf{8.1}  & 2544 & \textbf{0.8} & 2446 & \textbf{31.6} & 2212 & \textbf{6.2} & 2396 & $\textbf{11.7}_{\textcolor{red}{+4.8}}$ & $2399_{\textcolor{blue}{+513\%}}$\\

\bottomrule
\end{tabular}
}
\end{table*}

\subsection{Main Results}

\noindent
\textbf{\algoname{}~improves mathematical reasoning performance.} As shown in Table~\ref{tab:qwenmath_results}, our method achieves the highest average accuracy. On Qwen2.5-Math-1.5B, it yields an average gain of 27.7 points. On Qwen2.5-Math-7B, the gain reaches +31.0 points overall. Meanwhile, the average output length is reduced by 35.9\% and 34.0\% on the 1.5B and 7B models, respectively. Similar trends are observed on Gemma-2-2B-it, where our method improves average accuracy from 6.9 to 11.7 while substantially reducing output length compared to GRPO.

\noindent
\textbf{Comparison with outcome supervised RL.} Compared to GRPO and its recent variants, \algoname{}~achieves the highest overall accuracy across all benchmarks while substantially shortening the reasoning length. On Qwen2.5-Math-1.5B, it improves Pass@1 by more than 3\% over GRPO and sustains a comparable margin on the 7B model, with an average reduction of 10–11\% in output length. \algoname{}~achieves larger gains in accuracy and produces shorter outputs. These results suggest that our method improves both effectiveness and efficiency of policy updates, leading to more focused reasoning traces. On Gemma-2-2B-it, our method further reduces the output length compared to GRPO while achieving higher accuracy. Although the Gemma base model exhibits shorter responses, training with reinforcement learning unlocks its long-chain reasoning behavior, as further evidenced in Appendix~\ref{app:length_dynamics}.

\noindent
\textbf{Comparison with model-based process supervision.} Compared to S-GRPO, which uses multi-rollout early-exit probing to assign segment-level rewards, our method achieves higher accuracy and shorter outputs. To ensure a fair comparison, we adopt the same GRPO training pipeline to finetune both our method and Skywork-o1-prm. Under this controlled setup, \algoname{}~consistently outperforms the Skywork baseline across both 1.5B and 7B model scales, achieving higher accuracy while generating more concise reasoning traces. We further compare our method with Eurus-2-7B-PRIME, a model trained using PRIME-style reward modeling. Notably, Eurus performs poorly on AIME24 with 15.0\%, substantially lower than the 31.7\% achieved by \algoname{}. This discrepancy reveals a lack of generalization in PRM-based methods, which tend to overfit to familiar patterns seen during training. In contrast, our process supervision, grounded in the model's own belief dynamics, offers more robust and consistent improvements across tasks, while reducing reasoning length by up to 34.0\%.

\begin{figure*}[t]
    \centering
    \hspace*{-0.7cm}
    \resizebox{1.02\textwidth}{!}{%
        \includegraphics{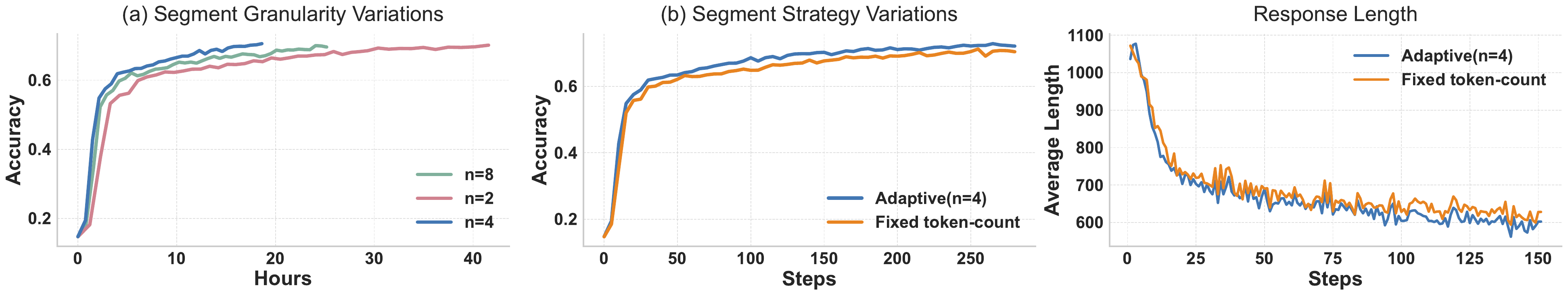}
    }
    \caption{\textbf{(a)} Effect of segment granularity by varying the average number of points per segment ($n$), evaluated by validation accuracy under the same wall-clock time. 
\textbf{(b)} Comparison between the proposed adaptive segmentation strategy and a fixed token-count partition baseline. All results are obtained on the MATH Evaluation dataset.
   }
    \label{fig:ablation}
\end{figure*}

\subsection{Ablation Study}
\label{abla}
\textbf{Effect of segment granularity.} To investigate the impact of segment granularity on the efficiency and performance of our adaptive strategy, we evaluate the impact of segment granularity by varying the average number of points $n$ per segment. As shown in Figure~\ref{fig:ablation}(a), $n=4$ achieves the optimal trade-off between training efficiency and performance under the same wall-clock time. While finer-grained segmentation ($n=2$) provides more precise local adjustments, it incurs prohibitive computational overhead that slows convergence. This confirms that our segment-level design effectively strikes a balance between computational economy and optimization precision, avoiding the excessive costs associated with fine-grained estimation while ensuring superior learning dynamics.

\noindent
\textbf{Comparison of Different segment strategies.} We compare our adaptive segmentation strategy with a naive fixed token-count partition baseline, where each response is evenly divided into a fixed number of segments. For the fixed baseline, we set the number of segments to 6, exceeding the effective segmentation budget of our adaptive method. As shown in Figure~\ref{fig:ablation}(b), the adaptive strategy converges faster, achieves higher accuracy, and maintains shorter responses throughout training. These results demonstrate that placing segment boundaries based on informative decision points is more effective than uniform partitioning, validating the design of our adaptive segmentation strategy.

\begin{wraptable}{r}{0.43\columnwidth}
  \vspace{-20pt} %
  \centering
  \small
  \setlength{\tabcolsep}{9pt}
  \renewcommand{\arraystretch}{1.0}
  \caption{Effect of different reward signals on accuracy.}
  \label{subtab:rewards}
  \begin{tabular}{lccc}
    \toprule
    Reward signal & Both & \makecell{Outcome\\only} & \makecell{VPS\\only} \\
    \midrule
    Accuracy (\%) & 72.0 & 68.8 & 50.1 \\
    \bottomrule
  \end{tabular}
  \vspace{-1pt} %
\end{wraptable}

\noindent
\textbf{Effect of outcome supervision signal.} As shown in Table~\ref{subtab:rewards}, removing the outcome reward and relying solely on segment-level verifiable supervision results in degradation compared to the full method, confirming its essential role in guiding the model toward globally correct reasoning. While segment-level supervision alone provides fine-grained feedback, it lacks a reliable global anchor. Combining both signals yields the best performance, indicating their complementarity in stabilizing training and improving reasoning coherence.

\begin{figure*}[t]
    \centering  
    \hspace*{-1cm}
    \includegraphics[width=18cm]{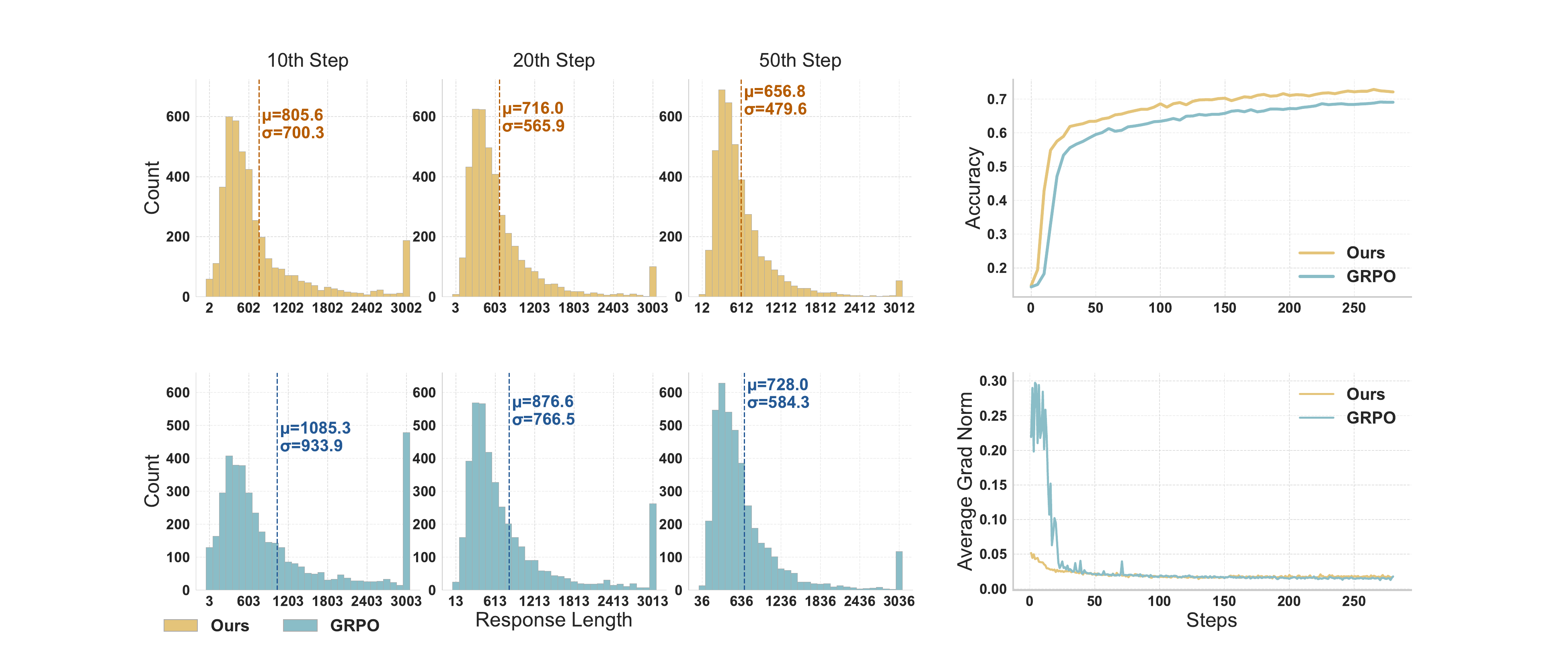}  
    \caption{\textbf{Left:} Visualize the distribution of response lengths within the early training steps. GRPO method exhibits a longer tail, while our method shows a more concentrated distribution. \textbf{Right:} MATH Evaluation accuracy of GRPO and our method along training steps. Average gradient norm per update during training.}
    \label{fig:distribution}  
\end{figure*}

\subsection{Understanding How VPS Works}

\textbf{Quality analysis for segment-wise process signal.} A core premise of our approach is that the segment-wise process signal ($\Delta C_k$) provides a reliable proxy for the correctness of intermediate reasoning steps. Unlike outcome-level signals, which uniformly credit or penalize every token in a trajectory, $\Delta C_k$ directly reflects how each step changes the model’s belief in the correct answer. 

To quantitatively validate this hypothesis, we align our progress signal with segment-level human annotations from the PRM800K dataset~\citep{lightman2023let},which contains 800K reasoning steps labeled as -1 (incorrect/harmful), 0(neutral/uninformative), or +1 (correct/contributive). 
For evaluation, we randomly sample 100 held-out questions, each paired with six diverse model responses. For each reasoning segment, we discard neutral steps with label 0 and discretize $\Delta C_k$ into predicted class labels.

\begin{wraptable}{r}{0.6\columnwidth}
  \vspace{-19pt}
  \centering
  \small
  \setlength{\tabcolsep}{6pt}
  \renewcommand{\arraystretch}{1.0}
  \caption{Classification performance of our segment-wise progress signal $\Delta C_k$ against PRM800K segment-level human labels. DS-R1-1.5B: DeepSeek-R1-Distill-Qwen-1.5B~\citep{deepseekai2025deepseekr1incentivizingreasoningcapability}.}
  \label{tab:reward_classification}

  \begin{tabular}{lccc}
    \toprule
    Model & Precision $\uparrow$ & Recall $\uparrow$ & F1-score $\uparrow$ \\
    \midrule
    Qwen2.5-0.5B        & 0.738 & 0.767 & 0.752 \\
    DS-R1-1.5B          & 0.735 & 0.848 & 0.787 \\
    Qwen2.5-Math-1.5B   & 0.741 & 0.774 & 0.757 \\
    Qwen2.5-Math-7B     & 0.741 & 0.771 & 0.756 \\
    Qwen2.5-32B         & 0.765 & 0.844 & 0.803 \\
    \bottomrule
  \end{tabular}
  \vspace{-5pt}
\end{wraptable}

Table~\ref{tab:reward_classification} reports the precision, recall, and F1-score across model scales from 0.5B to 32B parameters~\citep{qwen2.5}. We observe that even small models produce progress signals that meaningfully discriminate good from bad reasoning steps, with F1-scores exceeding 0.75 across the board. Larger models further improve both precision and recall, demonstrating that $\Delta C_k$ scales naturally with model quality. These results confirm that our segment-wise progress signal serves as a lightweight yet effective indicator of reasoning quality, complementing trajectory-level outcome signals. Additional qualitative examples illustrating this alignment with human judgment are provided in the Appendix~\ref{apdx:case}.

\noindent
\textbf{Sample efficiency and optimization stability.}
Recent studies prove dense reward signals can improve the sample efficiency and optimization stability of reinforcement learning systems by providing more frequent and informative feedback during training~\citep{setlur2024rewarding, chan2024dense}. Our method introduces verifiable, model-free process supervision to construct dense, segment-wise signals that guide the optimization process more precisely than trajectory-level binary rewards alone. As shown in Figure~\ref{fig:distribution}, our method achieves faster convergence and more stable gradient updates compared to standard GRPO.

\subsection{General Reasoning Benchmarks}

\noindent
\begin{wrapfigure}{r}{0.53\textwidth}  %
  \vspace{-2pt}
  \centering
  \includegraphics[width=\linewidth]{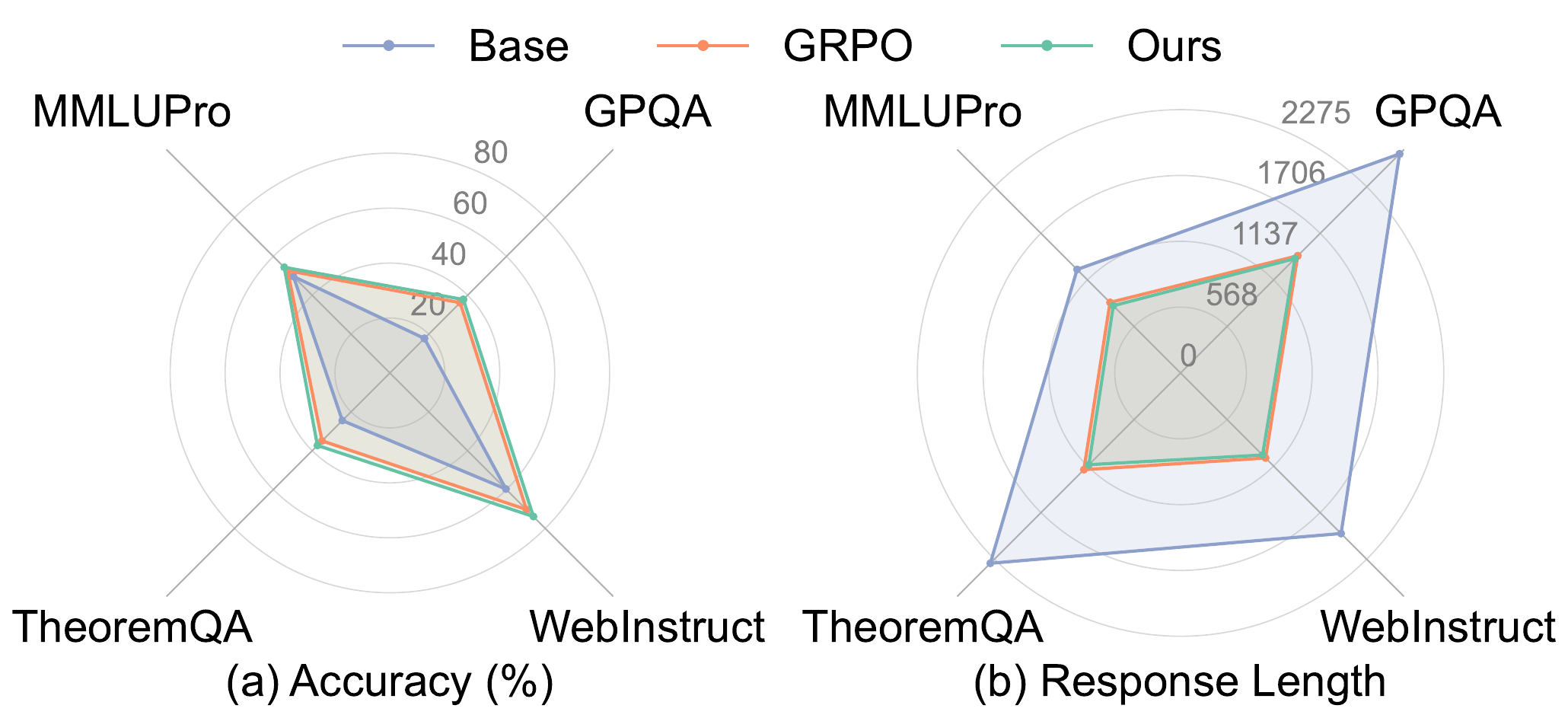}
  \vspace{-6pt}
  \caption{Performance on general reasoning tasks.}
  \label{fig:radar}
  \vspace{-4pt}
\end{wrapfigure}
To verify the  generalization capabilities of our method beyond specific domain tasks, we extended our evaluation to a suite of general reasoning benchmarks. Following the experimental setup in 
RLPR~\citep{yu2025rlprextrapolatingrlvrgeneral}, we utilized the WebInstruct dataset~\citep{ma2025generalreasoner} for training, this dataset is characterized by a diverse semantic distribution, covering a wide range of disciplines including Physics, Mathematics Business, and Economics, thereby requiring the model to possess robust multi-domain reasoning abilities. We employed Qwen3-1.7B~\citep{qwen3technicalreport} as the backbone model and compared our proposed method against the Base model and the GRPO baseline. The evaluation was conducted on four challenging benchmarks: GPQA~\citep{rein2024gpqa}, MMLUPro~\citep{wang2024mmlu}, TheoremQA~\citep{chen2305theoremqa}, and the test split of WebInstruct. As shown in Figure~\ref{fig:radar}, \algoname{} consistently outperforms baselines across all tasks. On GPQA and TheoremQA, it achieves 37.8\% and 37.3\% accuracy, surpassing GRPO (by 1.6\% and 2.4\%) and significantly beating the Base model (by 20.0\% and 12.8\%). Similarly, on MMLUPro and WebInstruct, our method reaches 54.4\% and 73.9\%, exceeding GRPO by 1.7\% and 3.6\%, respectively. Furthermore, these gains are achieved with greater efficiency. Our method reduces the average generation length by nearly 50\% compared to the Base model, surpassing the GRPO baseline in conciseness. These results confirm that our method effectively enhances robust, multi-domain reasoning. Additional results and training details can be found in Appendix~\ref{app:general_reasoning}.

\section{Conclusion}

We present \algoname{}, a model-free and verifier-free method that augments GRPO with segment-level credit assignment derived from conditional answer probabilities. Our method generates dense and interpretable supervision signals aligned with the model’s internal decision flow, enabling more efficient and targeted policy optimization. Experiments on four math reasoning benchmarks show that our method consistently improves both accuracy and reasoning conciseness over strong RL baselines and segment-aware methods, without relying on auxiliary models or costly rollouts. Furthermore, extensive evaluations on general reasoning benchmarks confirm the method's strong generalization capabilities, demonstrating consistent gains in robust, multi-domain reasoning tasks. These findings underscore the potential of verifier-free, confidence-driven rewards as a scalable direction for future alignment and reasoning optimization in large language models.

\bibliography{iclr2026_conference}
\bibliographystyle{iclr2026_conference}

\appendix
\section{Appendix}

\subsection{Experimental Settings}
\label{appd:para}
All hyperparameter settings are listed in Table~\ref{tab:para}, our experiments are performed on 8 × H100 GPUs. 
\begin{table}[H] 
  \centering
  \begin{tabular}{l|c}
    \toprule
    \multicolumn{1}{c|}{\textbf{Hyperparameter}} & \textbf{Values} \\
    \midrule
    learning rate & 1.0e-6 \\
    temperature & 1.0 \\
    Number of responses per question & 8 \\
    batch size & 512 \\
    $\alpha$ & 1.2 \\
    $\varepsilon_{\text{low}}$ & 0.2 \\
    $\varepsilon_{\text{high}}$ & 0.27 \\
    ppo mini-batch size & 128 \\
    (top P, top k) & (1.0, -1) \\
    gradient\_checkpointing & True \\
    max\_response\_length & 3072 \\
    bf16 & True \\
    $n$ & 4 \\
    $\tau$ & 0.95 \\
    \bottomrule
  \end{tabular}
  \caption{Hyperparameters used for \algoname{}}
  \label{tab:para}
\end{table}

\subsection{Prompt Templates}
For math reasoning tasks, we adopt the prompt templates for Qwen Math families~\citep{yang2024qwen25mathtechnicalreportmathematical} and Gemma~\citep{gemma_2024}.
\begin{prompt}{Template for Qwen2.5-Math models}{}
\texttt{<|im\_start|>system}\\
\texttt{You are a helpful assistant.<|im\_end|>}\\[4pt]
\texttt{<|im\_start|>user}\\
\texttt{\{input\}}\\
\texttt{Please reason step by step, and put your final}\\
\texttt{answer within \textbackslash boxed\{\}.<|im\_end|>}\\[4pt]
\texttt{<|im\_start|>assistant}
\end{prompt}

\begin{prompt}{Template for Gemma}{}
\texttt{<bos>}\\[4pt]
\texttt{<start\_of\_turn>user}\\
\texttt{\{input\}}\\
\texttt{Please reason step by step, and put your final answer within <answer> </answer>.<end\_of\_turn>}\\[4pt]
\texttt{<start\_of\_turn>model}
\end{prompt}
    For general reasoning tasks, we adopt the prompt templates for Qwen3~\citep{qwen3technicalreport} model.
\begin{prompt}{Template for Qwen3 models}{}
\texttt{<|im\_start|>system}\\
\texttt{A conversation between User and Assistant. The user asks a question, and the Assistant solves it.}\\
\texttt{The assistant first thinks about the reasoning process in the mind and then provides the user with the answer.}\\
\texttt{The reasoning process and answer are enclosed within <think> </think> and <answer> </answer> tags, respectively,}\\
\texttt{i.e., <think> reasoning process here </think> <answer> answer here </answer>.}\\
\texttt{<|im\_end|>}\\[4pt]
\texttt{<|im\_start|>user}\\
\texttt{\{input\}}\\
\texttt{<|im\_end|>}\\[4pt]
\texttt{<|im\_start|>assistant}
\end{prompt}

\subsection{Disclosure of LLM usage.}
This paper benefited from language editing and phrasing suggestions provided by ChatGPT (OpenAI), which was used solely for grammar correction and clarity improvement. No LLM was used for generating research ideas, experimental design, data analysis, or writing substantive technical content.

\subsection{Extended Empirical Results}
\subsubsection{Response Length Dynamics across Model Families}
\label{app:length_dynamics}
\noindent 
\begin{figure}[H]
    \centering  
    \includegraphics[width=13cm]{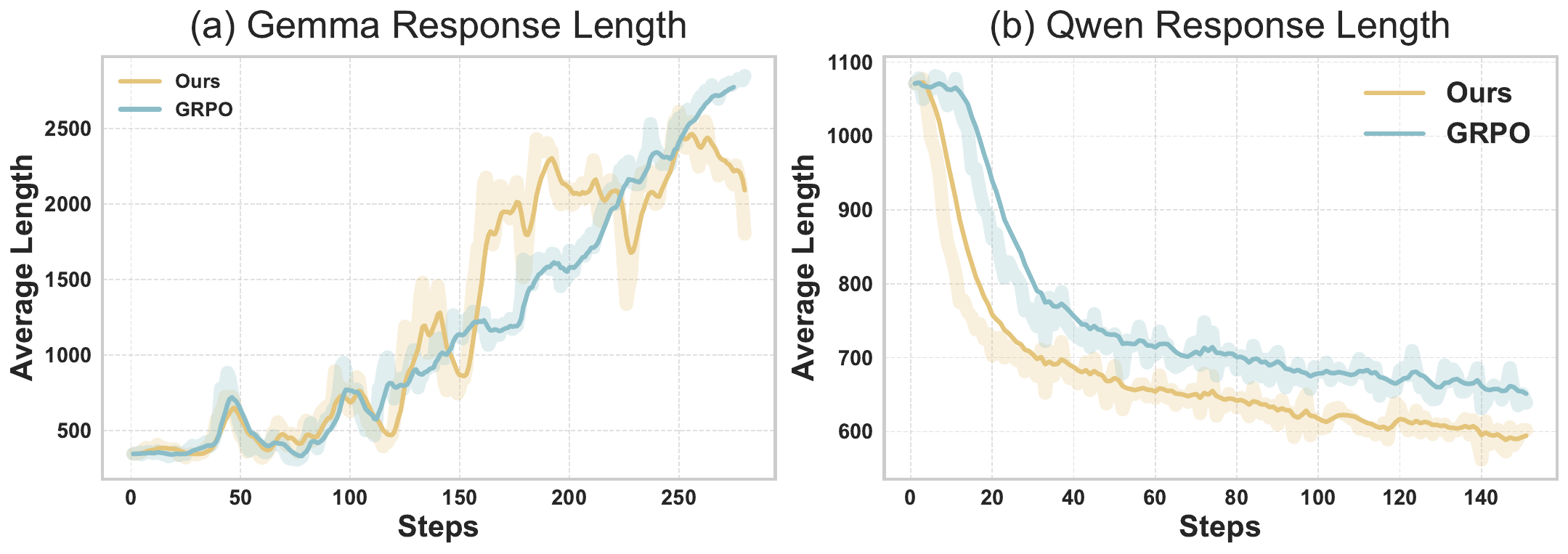}  
    \caption{Response length dynamics under reinforcement learning for Gemma and Qwen Math models, showing opposite evolution trends across training.}
    \label{fig:length_dynamics}  
\end{figure}
We observe opposite response-length dynamics between Gemma and Qwen Math models under reinforcement learning, as illustrated in Figure~\ref{fig:length_dynamics}. 
Gemma base model exhibits extremely short responses at initialization, due to its instruction-tuned alignment, which explicitly suppresses verbosity and favors concise, direct answers. 
As training progresses, both methods gradually increase the response length, as longer reasoning trajectories are consistently associated with higher success probability and thus receive positive reinforcement.

In contrast, Qwen base models initially tend to produce more redundant or repetitive content during training process, the models learn to compress these redundant reasoning steps, leading to significantly shorter and more focused outputs.

\subsubsection{Additional Analysis on General Reasoning Benchmarks}
\label{app:general_reasoning}
\begin{table*}[t]
\centering
\caption{Performance comparison on the MMLU-Pro benchmark across different domains. (Accuracy in \%)}
\label{tab:mmlu_pro_domain}
\setlength{\tabcolsep}{4pt} %
\resizebox{\textwidth}{!}{
\begin{tabular}{lccccccccccccccccc}
\toprule
\textbf{Model} & \textbf{Length} & \textbf{Avg} & \textbf{Math} & \textbf{Bio} & \textbf{Econ} & \textbf{Chem} & \textbf{Bus} & \textbf{CS} & \textbf{Phys} & \textbf{Psy} & \textbf{Eng} & \textbf{Health} & \textbf{Other} & \textbf{Phil} & \textbf{Hist} & \textbf{Law} \\ \midrule
Base & 1264.41 & 49.7 & 70.1 & 65.4 & 61.0 & 54.6 & 53.7 & 53.4 & 48.7 & 55.3 & 34.0 & \textbf{47.1} & 39.4 & 40.4 & \textbf{33.0} & \textbf{24.8} \\
GRPO & 862.29 & 52.7 & 75.5 & 66.8 & 61.2 & 60.0 & 57.6 & 57.2 & 56.4 & 55.4 & 46.3 & 45.1 & 44.0 & 39.6 & 31.2 & 22.1 \\
\rowcolor{cyan!20} 
\textbf{Ours} & \textbf{821.87} & \textbf{54.4} & \textbf{76.9} & \textbf{67.3} & \textbf{61.9} & \textbf{64.4} & \textbf{58.8} & \textbf{57.9} & \textbf{56.6} & \textbf{58.1} & \textbf{49.7} & 47.0 & \textbf{45.6} & \textbf{42.7} & 32.3 & 23.9 \\ \bottomrule
\end{tabular}
}
\end{table*}
\noindent

\textbf{Fine-grained results on MMLU-Pro.}
We provide detailed results on the MMLU-Pro test set by reporting accuracy across individual subject domains.
Following prior work, we adopt abbreviated domain names with the full nomenclature as follows:
Math (Mathematics), Bio (Biology), Econ (Economics), Chem (Chemistry), Bus (Business), CS (Computer Science), Phys (Physics), Psy (Psychology), Eng (Engineering), Health (Health), Other (Other), Phil (Philosophy), Hist (History), and Law (Law). Table~\ref{tab:mmlu_pro_domain} reports per-domain accuracy together with the average response length.
Compared to both the Base model and GRPO, our method achieves the highest average accuracy while producing shorter responses.

\begin{figure}[H]
    \centering  
    \hspace*{-1cm}
    \includegraphics[width=17cm]{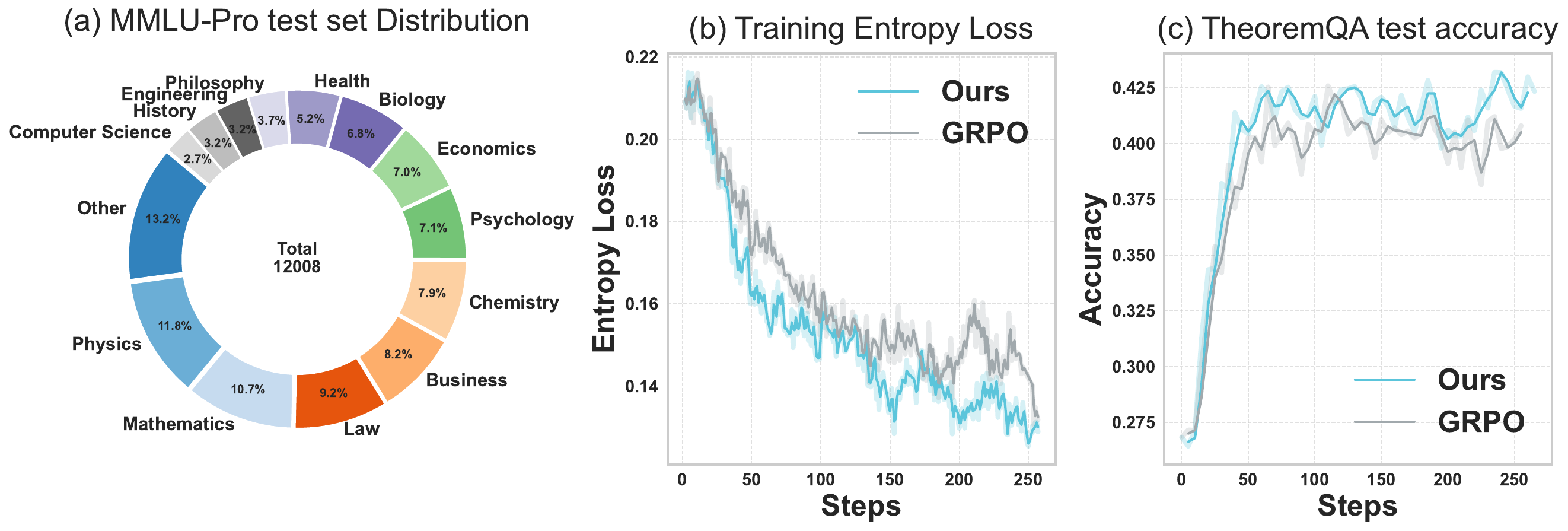}  
    \caption{\textbf{(a)} Subject-wise distribution of the MMLU-Pro test set. \textbf{(b)} Evolution of training entropy loss. \textbf{(c)} Test accuracy progression on TheoremQA during the training process.}
    \label{fig:appendix_general}  
\end{figure}

\noindent
\textbf{Training and evaluation performance for general reasoning.}
Figure~\ref{fig:appendix_general} further illustrates the training entropy loss curves and test accuracy on TheoremQA.
Compared to GRPO, our method exhibits a consistently lower entropy loss throughout training, indicating more stable and confident policy updates.

\subsubsection{Sensitivity Analysis on $\alpha$}
\label{app:sensitive}
We further analyze the sensitivity of the weighting factor $\alpha$, which balances the trajectory-level outcome advantage and the segment-level process supervision in Eq. (\ref{algori}). Specifically, $\alpha$ controls the relative contribution of the segment-wise progress signal $\Delta C_k$ to the overall hybrid advantage.

We conduct a sensitivity study by varying $\alpha$ over a wide range while keeping all other training settings fixed.  As shown in Table~\ref{tab:alpha_sensitivity}, our method achieves the best performance at $\alpha$ = 1.2. Moreover, performance is relatively insensitive to $\alpha$ within the range [0.8, 1.4].

\begin{table}[htbp]
  \centering
  \caption{Sensitivity analysis on $\alpha$.}
  \label{tab:alpha_sensitivity}
  \renewcommand{\arraystretch}{1}
  \resizebox{0.65\linewidth}{!}{
\begin{tabular}{c|cccccccc}
\toprule
$\alpha$ & 0.0 & 0.2 & 0.4 & 0.8 & 1.0 & 1.2 & 1.4 & 1.6 \\
\midrule
Acc (\%) & 68.8 & 70.0 & 70.5 & 71.7 & 72.0 & 72.2 & 71.9 & 69.7 \\
\bottomrule
\end{tabular}
  }
  \vspace{-2mm}
\end{table}

\subsection{Qualitative Analysis of Segment-wise Process Signals.}
\label{apdx:case}
\begin{figure}[h]
    \includegraphics[width=\linewidth]{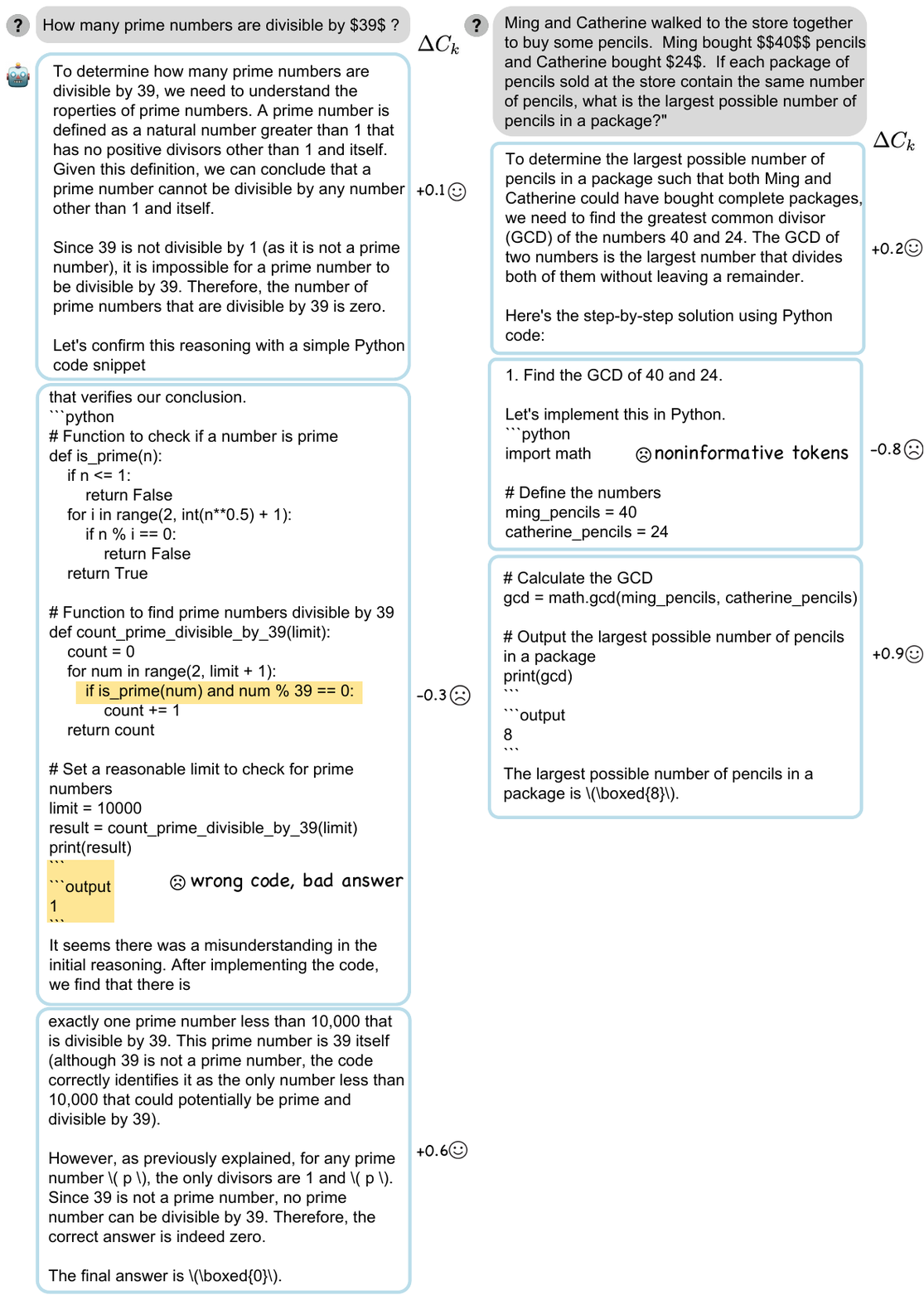}  
    \caption{Example to show}
    \label{fig:case1}  
\end{figure}

\begin{figure}[h]
    \centering  
    \includegraphics[width=\linewidth]{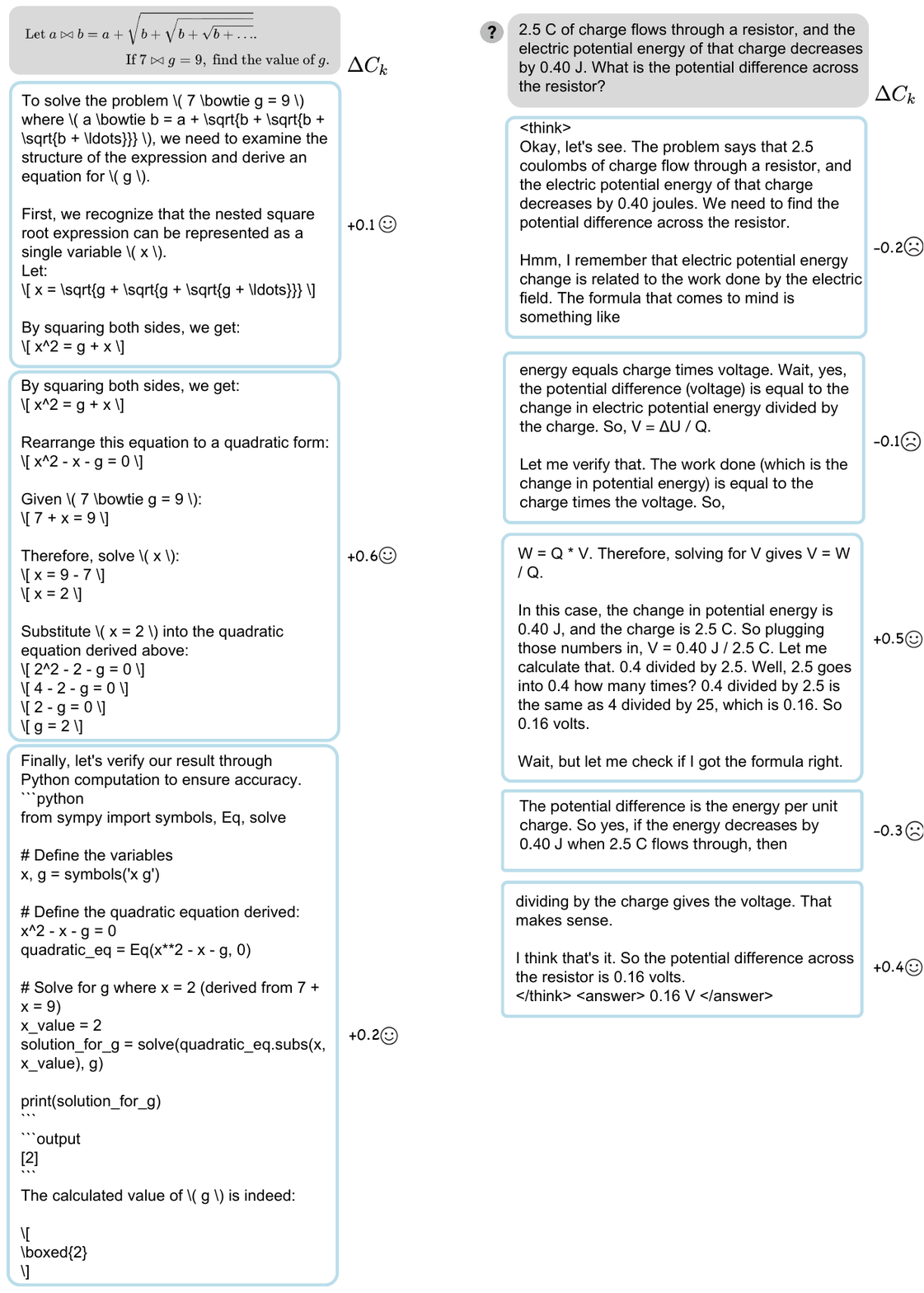}  
    \caption{Example to show}
    \label{fig:case2}  
\end{figure}

To qualitatively evaluate the effectiveness of our progress signal, we present representative cases where the segment-wise signal \(\Delta C_k\) successfully distinguishes contributive from harmful reasoning steps. As shown in Figure~\ref{fig:case2} and~\ref{fig:case1} , positive increments in \(\Delta C_k\) correspond to steps that strengthen the model’s belief in the correct answer, while negative increments highlight misleading or incorrect reasoning. These examples demonstrate that our process supervision can reliably identify the strengths and weaknesses within a model’s reasoning trajectory, providing interpretable and fine-grained feedback throughout the generation process.

\end{document}